\documentclass[12pt,journal]{article}
\usepackage{times,latexsym}
\usepackage{url}
\usepackage[T1]{fontenc}

\usepackage{amsmath}
\usepackage{amsfonts}
\usepackage{amssymb}
\usepackage{natbib}
\usepackage{algorithm}% http://ctan.org/pkg/algorithms
\usepackage{algpseudocode}% http://ctan.org/pkg/algorithmicx
\usepackage{amsthm}
\usepackage{dsfont}
\usepackage{tabularray}
\usepackage{bbm}
\usepackage{dsfont}
\usepackage{graphicx}
\usepackage{mathtools}
\usepackage[toc,page]{appendix}
\usepackage{float}

\numberwithin{equation}{section}
\DeclareMathOperator*{\argmax}{arg\,max}
\DeclareMathOperator*{\argmin}{arg\,min}

\begin{document}
\date{}
\title{
Transferring Neural Potentials For High Order Dependency Parsing
}
\author{Farshad Noravesh\footnote{Email: noraveshfarshad@gmail.com}}
\maketitle
%	  \large\textit{Farshad Noravesh}\footnote{Email: noraveshfarshad@gmail.com}
\begin{abstract}
High order dependency parsing leverages high order features such as siblings or grandchildren to improve state of the art accuracy of current first order dependency parsers. The present paper uses biaffine scores to provide an estimate of the arc scores and is then propagated into a graphical model. The inference inside the graphical model is solved using dual decomposition. The present algorithm propagates biaffine neural scores to the graphical model and by leveraging dual decomposition inference, the overall circuit is trained end-to-end to transfer first order informations to the high order informations. 
%Keywords:  
\end{abstract}

\section{Introduction}
Dependency parsing is the basis of many complex pipelines for problems in natural language processing such as machine summarization, machine translation, event extraction, semantic parsing ,semantic role labeling(SRL), emotion analysis, dialogue systems and information processing. Thus, any error in dependency parsing could propagate to downstream task and therefore any advance in this field could lead to major improvement in NLP tasks.
There are two main approaches to dependency parsing. 

The first approach is transition based which has incremental local inference and involves using datastructures such as buffer and stack \citep{Nivre2008},\citep{Buys2015}. This approach has the limitation of resolving relatively short sentences and is a trade-off between speed and accuracy.

The second approach is graph based and can handle any long sentence but the inference time has usually long time complexity.

There are many technical issues  for improving state of the art dependency parsers. Thus these research directions include: \\

 \begin{enumerate}
    \itemsep0em 
    \item nonprojective cases
    \item high order features
    \item faster inference algorithms
    \item training data scarcity and the need for few shot learning
    \item span-span modeling
    \item reranking
\end{enumerate}
The amount of nonprojective examples of dependency parsing varies from one language to another. At inference time, Eisner algorithm which is a bottom up CKY-like algorithm can not resolve nonprojective cases. For nonprojective parsing, many algorithms based on maximum spanning tree are used such as \citep{Zmigrod2021},\citep{McDonald2005},\citep{McDonald2006},\citep{Levi2015} or leveraging maximum subgraph parsing for the case of nontrees in semantic dependency parsing as is shown in \citep{Kuhlmann2015}.
An alternative approach to finding maximum spanning tree is to formulate it as an integer linear programming(ILP) like \citep{Riedel2006}.
One advantage of ILP formulation is the capability to model many constraint like each verb has a single subject in a direct way. The other advantage is that it can be used for nonprojective cases \citep{RiedelAndClarke2006}.

Most articles in the literature are devoted to first order parsing \citep{Dozat2016} which only use one edge and node as sources of input features and neglect the richness of language structure. Although \citep{Li2022} uses graph neural networks(GNN) and creates the graph dynamically, but it still does not model high order features. A good way to leverage GNN to consider higher order features like grandparents, grandchildren and siblings, is described in \citep{Ji2019} which recursively aggregates the neighbors’ information and the graph at each layer appears as a soft parse tree, and finally it uses a MST algorithm for decoding. The novelty of this GNN is that it represents both the head representation of each node, and dependent representation of each node. The drawback is that the number of parameters to learn is quite large and it also suffers from the curse of dimensionality and therefore needs many data to train efficiently in this high dimensional vector space. An interesting idea to get around this difficulty in GNN, is considering each node as an edge in the dependency structure, which is explained in \citep{Yang2022}. An alternative for GNN is stack-pointer networks in \citep{Ma2018} and siblings and grandchildren features have been modeled. Although this model is fast but has all the limitations of deep learning such as interpretability, being data hungry and struggling with the curse of dimensionality.

One way to generalize to high order features is using ILP and see it as a structural prediction problem \citep{MartinsAndSmith2009}. In order to unify all approaches, graphical models are used as a promising paradigm as is shown in \citep{Niculae2020}. Prior knowledge could be encoded as hard constraint \citep{Martins2009} and keeps polynomial number of constraints in general. \citep{Martins2009} uses nonlocal(high order) features. Another idea that can be combined is dual decomposition which is inspired by optimization \citep{Martins2011},\citep{MartinsAndFigueiredor2011}. A good approach to unify the loopy belief propagation parser of \citep{Smith2008} and the relaxed linear program \citep{Martins2009} is explained in \citep{Martins2010} that considers model assumptions in a factor graph. \citep{Matthew2015} considers Feed-forward topology of inference as a differentiable circuit and considers high order interactions in a factor graph. \citep{Matthew2015} models each potential function as a loglinear form.

Although high order features are crucial to obtain state of the art models for dependency parsing but there is another factor which is even more important and is described in \citep{Gan2021}. The basic idea is to measure the relation between the spans in contrast to measuring the relations between words in classical dependency parsing. This approach is a proper generalization since each word is always a span of length one, and subspans could be evaluated from spans recursively which could be considered as a dynamic programming paradigm.
\section{Related Works}
An inspiring and natural approach to high order dependency parsing is described in the seminal work of \citep{Smith2008}  that formulates it as an approximate learning and inference over a graphical model and the global constraints are encoded inside the model and Loopy Belief Propagation(LBP) is a simple approximation that is used for that. \citep{Smith2008} incrementally adjusts the numerical edge weights that are fed to a fast first-order parser. One of the main difficulties is satisfy hard constraints such as tree constraint which ensures the resulting graph is a tree. 
The probability distribution of all configurations(all assigments $\mathcal{A}$) is defined by the following Markov random field(MRF)
\begin{equation}\label{distribution}
p(\mathcal{A}) = \frac{1}{\mathcal{Z}}\underset{m}\prod F_{m}(\mathcal{A})
\end{equation}
where $F_{m}$ is the m-th factor function which could be unary, binary, ternary, or global. From a different classification angle, these factors could be either hard or soft. A hard factor has a value 0 on violated constraint parses, acting as a constraint to rule them out such as TREE constraint which ensures the final graph is a tree or even a harder constraint such as PTREE which ensures trees to be projective. Another important hard constraint is EXACTLY1 which does not allow any word to have more than one parent. Soft factors in \eqref{distribution} can easily be modeled by the following loglinear model:
\begin{equation}\label{factorm}
F_{m}(\mathcal{A}) = \exp \sum_{h\in features(F_{m})} \theta_{h}f_{h}(\mathcal{A},W,m)
\end{equation}
Nine types of soft constraints and seven hard constraints are described in \citep{Smith2008}. An interesting experimental and combinatorial exploration is discovering which set of soft and hard constraints are sufficient for a reasonable accuracy which experimentally measures the degree of sensitivity of each of these constraints to the final accuracy.

The main difficulty in training this model is the fact that the normalizing constant in the denominator depends implicitly on the learning parameters and therefore can not be neglected but Belief Propagation(BP) provides an estimate of that marginal distribution. Thus the gradient of the normalizing constant can easily be computed as follows.
\begin{equation}
\nabla_{\theta} \log \mathcal{Z} = \sum_{m} \mathbb{E}_{p( \mathcal{A} )} [ \nabla_{\theta} F_{m}(\mathcal{A})]
\end{equation}

\citep{Matthew2015} considers approximations and parsing in \citep{Smith2008} as a differentiable circuit to improve accuracy.
It uses a different objective function which is based on the $L2$ distance between the approximate marginals and the gold marginals.

\section{Main Results}
\subsection{Terminology}
Let $W = W_{0},\ldots,W_{n}$ denote the input sentence where $W_{0}$ is the root. The corresponding part of speech(POS) tags are $T_{1},\ldots,T_{n}$. There are $O(n^{2})$ links in the dependency parse that can be enumerated by $\{L_{ij}: 0\le i \le n , 1\le j\le n\}$

\subsection{Transferring Neural Potentials}
By borrowing from \citep{Dozat2016}, the scores can easily be calculated as follows:
\begin{equation}\label{biaffine-scores}
\begin{split}
h_{i}^{(arc-dep)}&=MLP^{(arc-dep)}(r_{i})    \\
h_{j}^{(arc-head)}&= MLP^{(arc-head)}(r_{j})     \\
s_{i}^{(arc)}&= H^{(arc-head)}U^{(1)}h_{i}^{(arc-dep)}+H^{(arc-head)}u^{(2)}
\end{split}
\end{equation}

Unary and binary potentials could be defined as follows:
\begin{equation} \label{eq:neural-potentials}
\begin{split}
\psi_{Y_{k}} &= \exp{s_{i(k)j(k)}} \\ 
\psi_{Y_{k},Y_{k'}} &= \psi_{Y_{k}}+\psi_{Y_{k'}}+ \phi_{Y_{k},Y_{k'}}
\end{split}
\end{equation}
where $i(k)$ and $j(k)$ is the simple lookup table mapping from the actual dependency graph to the graphical model. Please note that the score of the labels are also defined similarly.
The best way to understand the idea of transferring neural potentials is to imagine that the cheap and fast first order parser is the baseline and the goal is to perturb these edge scores to be adjusted to the global constraints through high order features. There are two paradigms that can resolve this issue. The first paradigm says that the weights of the first order parser does not receive any feedback from high order features and the misalignment is modeled by a new term $\phi_{Y_{k},Y_{k'}}$ that is small only for cases that high order features do not have any conflict with first order features and we call it a perfect alignment case.  
The second paradigm couples first order with high order features in a bidirectional way and allows to change the scores of first order parser by end to end training and the error is propagated all the way downstream to influence and tune the first order parser. The first paradigm can be used as a warm start of the second paradigm to increase the speed of training process since the initial weights are at a reasonable space and just a perturbation of it could satisfy the high order dependency constraints. The present paper assumes that $\psi_{Y_{k}}+\psi_{Y_{k'}}$ can sufficiently model the interactions of edges and there is no need to model the mutual interaction $\phi_{Y_{k},Y_{k'}}$ explicitly, since the model is trained end to end and all potentials are based on neural networks and the mutual interaction is implicitly considered.
There are two main approaches to inference for the best parse. The first one is based on sum-product algorithm also known as belief propagation algorithm. After calculating the beliefs from final message passing iteration, the marginal probability can be approximated. This should be done for all variable nodes of the factor graph to get all parts of the parse. The second approach simultaneously maximizes the objective by finding the best assignment which is also called the MAP assignment task. The second approach is mathematically richer since the integrality gap can be evaluated in contrast to loopy belief propagation that only can hope to reach the convergence and the evaluation is hard. These two approaches are explained here:
\subsubsection{Loopy Belief Propagation}
After sending messages iteratively from variables, $y_{i}$, to factors, $\alpha$ and, from factors to variables, the algorithm will eventually converge:
\begin{equation}
\begin{split}
m^{(t)}_{i\rightarrow  \alpha}(y_{i}) &\propto \prod_{\beta \in \mathcal{N}(i) \backslash \alpha} m_{\beta\rightarrow i}^{(t-1)}(y_{i})  \\
m^{(t)}_{\alpha\rightarrow  i}(y_{i}) &\propto \sum_{y_{\alpha}\sim y_{i}} \psi_{\alpha}(y_{\alpha}) \prod_{j \in \mathcal{N}(\alpha)\backslash i } m_{j\rightarrow \alpha}^{(t-1)}(y_{i})
\end{split}
\end{equation}
 where $\mathcal{N}(i)$ and $\mathcal{N}(\alpha)$ are the neighbors of $y_{i}$ and $alpha$ respectively. Beliefs at each variable and factor are computed as follows:
 \begin{equation}
\begin{split}
b_{i}(y_{i})&\propto \prod_{\alpha \in \mathcal{N}(i)} m_{\alpha\rightarrow i}^{(t_{max})}(y_{i}) \\
b_{\alpha}(y_{\alpha})&\propto \psi_{\alpha}(y_{\alpha}) \prod_{i\in \mathcal{N}(\alpha)} m_{i\rightarrow \alpha}^{(t_{max})} (y_{i})
\end{split}
\end{equation}
This approach is used in  \citep{Matthew2015} in the inference step.  
\subsubsection{MAP Inference}
In the present work, this approach is chosen since it is fast, parallelizable  and has a rich mathematical analysis.
Linear programming(LP) relaxation is used to solve MAP inference as is explained in \citep{Jaakkola2010}. The factor graph has an equivalent Markov random field(MRF) and thus the objective is as follows:
\begin{equation}
\begin{split}
\text{MAP}(\theta)&=\underset{\mu}{\max}\sum_{i\in V}\sum_{x_{i}}\theta_{i}(x_{i})\mu_{i}(x_{i})+\sum_{ij\in E} \sum_{x_{i},x_{j}}\theta_{ij}(x_{i},x_{j})\mu_{ij}(x_{i},x_{j}) \\
&=\underset{\mu }{\max}\ \theta . \mu
\end{split}
\end{equation}
subject to :  
\begin{equation} \label{eq-polytope}
\begin{split}
\mu_{i}(x_{i}) &\in \{0,1\}  \ \forall i\in V , x_{i}   \\
\underset{x_{i}}{\sum} \mu_{i}(x_{i}) &=1 \ \forall i\in V   \\
\mu_{i}(x_{i})&=\underset{x_{j}}{\sum}\mu_{ij}(x_{i},x_{j})  \ \forall ij\in E ,x_{i}         \\
\mu_{j}(x_{j})&=\underset{x_{i}}{\sum}\mu_{ij}(x_{i},x_{j}) \ \forall ij\in E ,x_{j}
\end{split}
\end{equation}
where $\theta_{i}$ and $\theta_{ij}$ are unary and binary potentials respectively.
This is a pairwise relaxation. We can tighten the relaxation by enforcing the joint consistency of edges in a cluster of variables using the framework of lift-and-project methods but is out of the scope of the present paper since a fast algorithm is more preferred to a highly accurate algorithm.  The lifting refers to introducing the new high level variables and the projection refers to projecting to the original variables. An alternative framework is to use cutting plane algorithms. When using these higher order methods, the number of constraints and the variables, grows exponentially in the size of the clusters considered and is therefore prohibitive.  
The constraints in \eqref{eq-polytope} can be generalized to the cluster based constraint as is done in \citep{Batra2011},\citep{Sontag2008} to have tighter relaxation. 
A different LP relaxation for MAP assignment problem is by reducing it to an instance of a Bipartite Multi-cut problem as is shown in \citep{Reddi2010}. A good survey of all LP relaxations for MAP inference in discrete Markov random fields is described in \citep{Kannan2019}.
A cutting-plane algorithm is used in the present paper as follows: After solving the pairwise LP relaxation, there are two cases. The first case is that the solution is integer, the MAP assignment is done, and algorithm is terminated.  To handle the second case, one can add a valid constraint to the relaxation. Valid constraint is a constraint that does not cut off any of the integral vertices.
Solving \eqref{eq-polytope} is computationally expensive and is not efficient. Thus, a natural approach is to use dual decomposition which is explained in \citep{MartinsAndFigueiredor2011}, \citep{Koo2010},\citep{Martins2011}.
Block coordinate descent is used for dual decomposition in \citep{Belanger2014} while the present paper used ADMM as is leveraged in \citep{Martins2011}.
\subsubsection{Dual Decomposition}
Following the ADMM approach of \citep{Martins2011} to dual decomposition, first the primal problem is defined as follows:
\begin{equation} \label{eq:consistency}
\begin{split}
P:  \underset {z_{s} \in y_{s}}{\max} \sum_{s=1}^{S} f_{s}(z_{s}) \\
<u(r)>_{r\in R} \in \mathbb{R}^{|R|} \\
s.t. \  z_{s}(r)=u(r) \  \forall s,r\in \bar{R}_{s}
\end{split}
\end{equation}
After doing relaxation and writing the dual form, the master problem is:
\begin{equation}
\begin{split}
D: \underset{\lambda=<\lambda_{1},\ldots,\lambda_{S}>}{\min} \sum_{s=1}^{S}g_{s}(\lambda_{s}) \\
s.t. \sum_{s:r\in \bar{R_{s}}} \lambda_{s}(r)=0 \  \forall r\in R
\end{split}
\end{equation}
where $g_{s}(\lambda_{s})$ are the solution to the following slaves
\begin{equation}\label{eq-max}
\underset{z_{s}\in Z_{s}}{\max} \  f_{s}(z_{s})+ \sum_{r\in\bar{R_{s}}} \lambda_{s}(r)z_{s}(r)-\frac{\rho}{2}\sum_{r\in \bar{R}_{s}} (z_{s}(r)-u^{t}(r))^{2}
\end{equation}
Since the scores $f_{s}(z_{s})$ in \eqref{eq-max} is modeled by a linear form like $f_{s}(z_{s})=\sum_{r\in R_{s}}\theta_{s}(r)z_{s}(r)$, the slaves can be written as:
\begin{equation}\label{eq-max-final}
\underset{z_{s}\in Z_{s}}{\max} \   \sum_{r\in\bar{R_{s}}} (\theta_{s}(r) +\lambda_{s}(r))z_{s}(r)-\frac{\rho}{2}\sum_{r\in \bar{R}_{s}} (z_{s}(r)-u^{t}(r))^{2}
\end{equation}
Note that $\theta_{s}(r)$ in \eqref{eq-max-final} are neural potentials that are estimated from the deep learning module and these coefficients vary at each iteration of the overall circuit.
To solve \eqref{eq-max-final} using a generic quadratic solver, it is written in the following form:
 \begin{equation}\label{eq:slave}
 \underset{z_{s}\in Z_{s}}{\max} \   \sum_{r\in\bar{R_{s}}} (\theta_{s}(r) +\lambda_{s}(r) +\rho u^{t}(r))z_{s}(r)-\frac{\rho}{2}\sum_{r\in \bar{R}_{s}} (z_{s}(r))^{2}
 \end{equation}
The Lagrange variables can be updated as follows:
\begin{equation}\label{eq:landa-update}
\lambda_{s}^{t+1}(r)=\lambda_{s}^{t}(r)-\eta_{t}(z_{s}^{t+1}(r)-u^{t+1}(r))
\end{equation}
where $\eta_{t}$ is the step size.
Applying ADMM algorithm, u has a closed form solution as a simple average which is obtained by projected subgradient method:
\begin{equation} \label{eq:u-update}
u^{t+1}(r)=\frac{1}{\delta(r)} \sum_{s:r\in \bar{R_{s}}} z_{s}^{t+1}(r)
\end{equation} 
where $\delta(r)$ is the cardinality of set $\{s: r \in R_{s}\}$.
The loop iterates until primal and dual residuals defined in \citep{Martins2011} violate the constraint. 
To fully perceive the details of these variables, consider a sentence with 5 tokens as follows:
$w_{1},\ldots,w_{5}$
\begin{figure}[H]
\centering
\includegraphics[scale=.4]{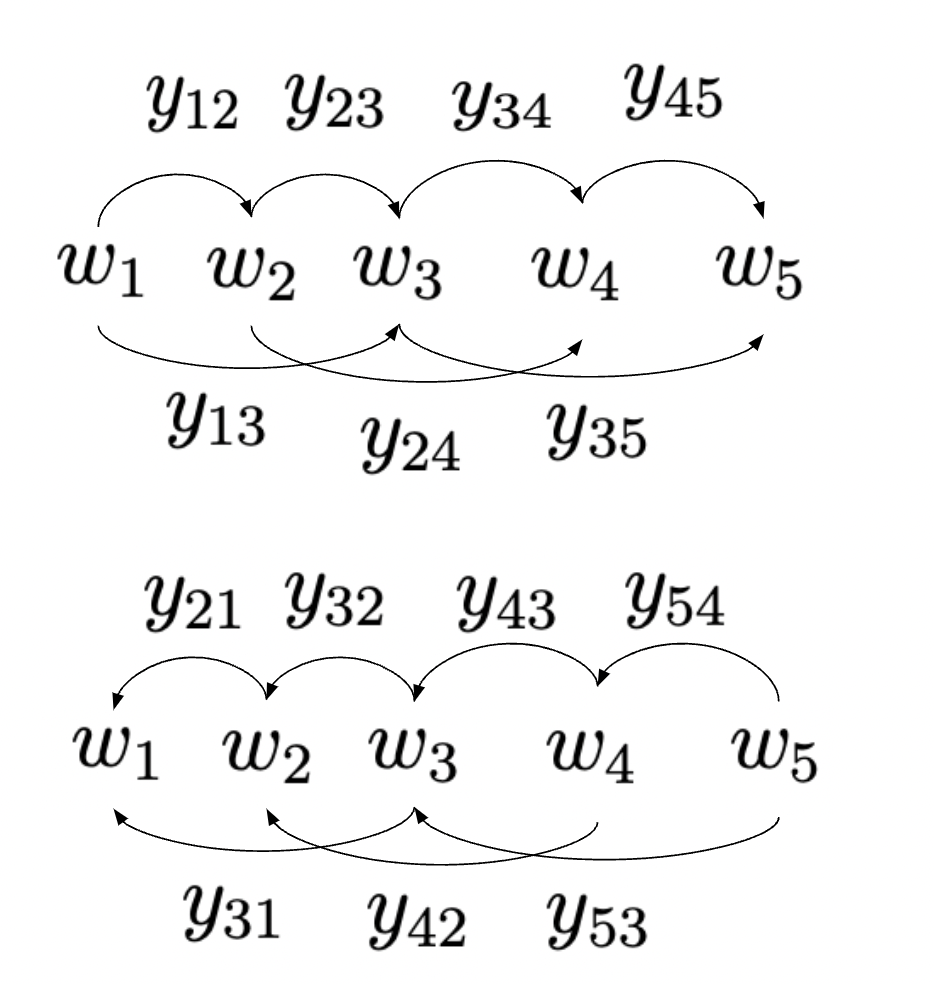}
\caption{right and left minimal dependencies}
\label{fig:minimaldep}
\end{figure}
$z^{gp}_{3f}=\{y_{34},y_{45}\}$

\begin{figure}[H]
\centering
\includegraphics[scale=.4]{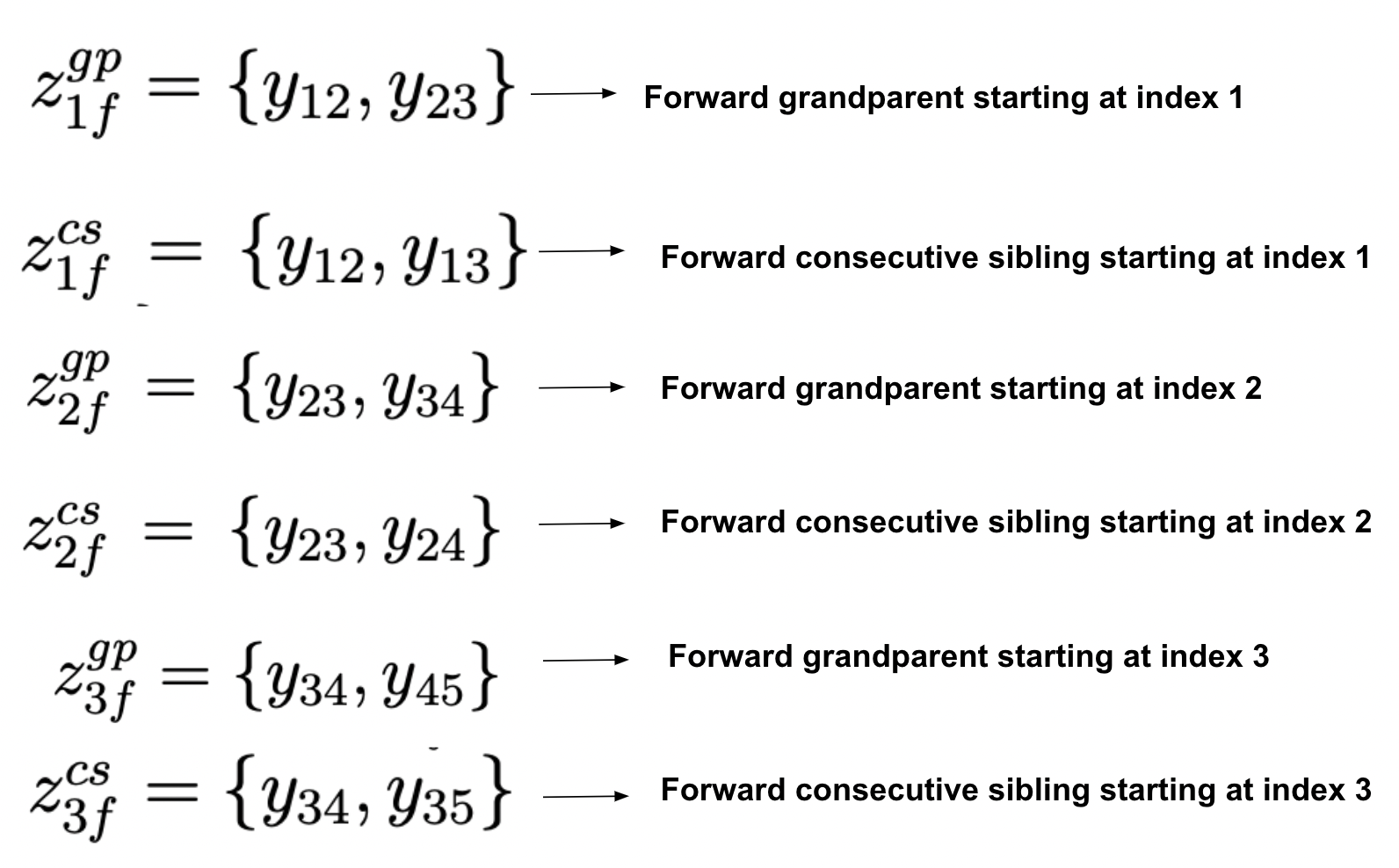}
\caption{forward constraints}
\label{fig:forwardConstraints}
\end{figure}

\begin{figure}[H]
\centering
\includegraphics[scale=.4]{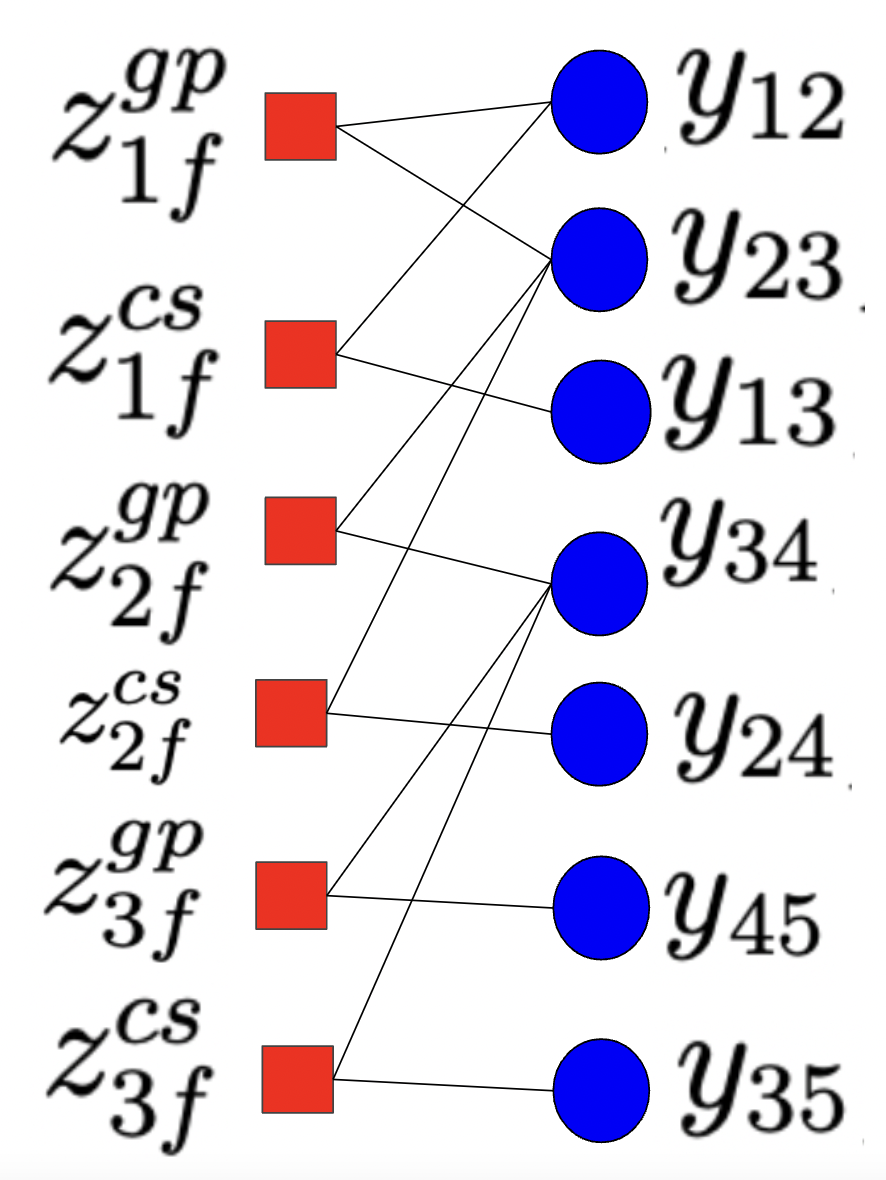}
\caption{factor graph}
\label{fig:factorGraph}
\end{figure}

The minimal dependencies are defined as all second order dependencies that are either consecutive siblings or grand parents and are shown in Figure~\ref{fig:minimaldep} with constraints that are shown in  Figure~\ref{fig:forwardConstraints} .
Figure~\ref{fig:factorGraph} shows the factor graph generated for 5 token sentences which includes 6 constraints and 7 overlapping basic components. Nots that only two types of higher order constraints namely grandparent and consecutive siblings are used in the present paper since there is always a tradeoff between computational complexity and exactness of the solutions. These two types of constraints are more essential than the rest of them and have more impact in any selection process.
In order to connect the dual decomposition to the first order deep learning model, a mapping is defined that assigns a score to each of the components. The weighted combinations of these first order scores supports how solution of the $z_{s}$ variables can shape the global dependency parsing graph:
\begin{equation}
f_{s}(z_{s})=\sum_{r} z_{(s,r)} \theta(s,r)
\end{equation}
Once the $z$ variables are solved, the dependency graph can be read. Note that the number of components of each slave is fixed but each basic component can be connected to any number of constraints. The equality constraint in equation~\eqref{eq:consistency} ensures the consistency of the selection.
\subsection{Training And Prediction}
By drawing inspiration from \citep{Matthew2015}, everything is trained end to end like a circuit and the inference mechanism of the factor graph is coupled with the neural estimation of edge scores. A drawback of the present approach like any other deep learning model is the lack of sufficient training data since deep learning models are data hungry and suffer from curse of dimensionality. There is a compromise between speed of convergence and the accuracy of modeling. This is the reason that very few slaves(constraints) are considered for high order modeling. The more constraint added to the model, the longer it takes to converge. 
Minimum Bayes risk(MBR) is used to produce a tree that minimizes the expected loss as follows:
\begin{equation}\label{loss}
\begin{split}
h_{\theta}(x)&=\underset{\hat{y}}{\argmin} \mathbb{E}_{y\sim p_{\theta}(.|x)} l(\hat{y},y) \\
 &=\underset{\hat{y}}{\argmax} \sum_{i:\hat{y_{i}}=ON} p_{\theta}(y_{i}=ON|x)
\end{split}
\end{equation}

At test time, maximum spanning tree automatically ensures that the resulting graph is a tree. 

\begin{algorithm}[H]
\caption{outputs dependency parsing tree}
\label{alg:dependencyParsing}
1:\textbf{Input}: batch of sentences \\
2:\textbf{repeat} \\
3: \ calculate word embedding \\
4: \ calculate the scores using \eqref{biaffine-scores} \\
5: \ calculate the neural potentials using first order estimation \eqref{eq:neural-potentials} \\
6: \  \textbf{repeat} \\
7: \ \ \ \textbf{for each} slave $s=1,\ldots,S$ do \\
8:  \  \ \ \ \ make an update for slave $z_{s}$ using \eqref{eq:slave}  \\
9:  \ \ \  \textbf{end for} \\
10: \ \ \ update u using \eqref{eq:u-update} \\
11: \ \ \ update $\lambda$ using \eqref{eq:landa-update} \\
12: \ \ \ $t \leftarrow t+1$ \\
13: \ until primal and dual residuals are below a threshold \\
14: \ round $u,z,\lambda $ \\
15: \ backpropagate the loss in \eqref{loss} to adjust the neural scores \\
16: until maxIter \\
17: \textbf{if not} tree \\
18: \ \ use maximum spanning tree algorithm \\
19: return the dependency parsing tree

\end{algorithm}

\subsection{Experiments}
Universal dependency dataset for English language is used for all of the experiments in the present paper. The maximum length of the sentence is 71 tokens but at most 60 token sentences are used for training since few datapoints for longer sentences are not enough for an adequate training and makes some noise.
\begin{table}[h!]
\begin{center}
	\caption{different experiments for dependency parsing}
   	\label{fig:experiments}
        \begin{tabular}{ |c | c | c | c | c | c | c | }
        \hline
	{experiment} & {opt} & {epochs}& {batch size} &{maxseq} &{highorder} & {accuracy}  \\
        \hline
            1 & Adam & 10 & 5 & 20 & False & 93.2 \\
      	2 & Adam & 10 & 5 & 40 & False & 91.2 \\
	3 & SGD & 10 & 5 & 60 & False & 90.1\\
	4 & Adam & 10 & 20 & 60 & False & 89.7 \\
	5 & Adam & 10 & 5 & 60 &False & 88.6 \\
	6 & Adam & 10 & 5 & 60 &True & 65.3 \\
	7 & SGD & 10 & 5 & 60 &True & 66.2 \\
        \hline
        \end{tabular}
\end{center}
\end{table}
When highorder in Table~ref{fig:experiments} is True, it means that the high order dependency parsing is used while when it is False, it corresponds to the first order modeling that uses biaffine attention and factor graph and dual decomposition algorithm are not involved in training.
One reason that the results of highorder in Table~\ref{fig:experiments} are lower than the first order counterpart, is that very few slaves are used in these experiments for faster convergence. Even the backward constraint slaves are not used and there is always a tradeoff between speed and accuracy.

 \section{Conclusion}
The present algorithm combines first order information with high order information to represent a richer feature representation to obtain relations and labels for dependency parsing. The contribution of the present paper is on joining deep biaffine scores in traditional deep learning with high order constraints that are best represented by graphical models. By combining the strength of deep learning and graphical model inferences, a unified approach to high order parsing is suggested. One can also analyze that the high order parser is a perturbation of first order parser. The analysis could also be used for further investigation on which structures in languages have high deviations between first order and high order parsers.

\bibliographystyle{agsm}
\bibliography{neuralpotential}
%\begin{appendices}
%\section{Appendix}

%\end{appendices}
\end{document}